\documentclass[a4paper,twoside]{article}
\usepackage{epsfig}
\usepackage{subcaption}
\usepackage{calc}
\usepackage{amssymb}
\usepackage{amstext}
\usepackage{amsmath}
\usepackage{amsthm}
\usepackage{multicol}
\usepackage{pslatex}
\usepackage{apalike}
\usepackage{bm}
\usepackage[bottom]{footmisc}
\usepackage{SCITEPRESS}     
\usepackage{framed,multirow}

\begin{document}

\title{Classifying Soccer Ball-on-Goal Position Through Kicker Shooting Action}

\author{\authorname{Javier Torón Artiles \orcidAuthor{0009-0000-5082-310X},
Daniel Hernández-Sosa\orcidAuthor{0000-0003-3022-7698}, Oliverio J. Santana\orcidAuthor{0000-0001-7511-5783}, Javier Lorenzo-Navarro\orcidAuthor{0000-0002-2834-2067}, and
David Freire-Obregón\orcidAuthor{0000-0003-2378-4277}}
\affiliation{SIANI, Universidad de Las Palmas de Gran Canaria, Las Palmas de Gran Canaria, Spain}
\email{david.freire@ulpgc.es}
}


\abstract{
This research addresses whether the ball's direction after a soccer free-kick can be accurately predicted solely by observing the shooter's kicking technique. To investigate this, we meticulously curated a dataset of soccer players executing free kicks and conducted manual temporal segmentation to identify the moment of the kick precisely. Our approach involves utilizing neural networks to develop a model that integrates Human Action Recognition (HAR) embeddings with contextual information, predicting the ball-on-goal position (BoGP) based on two temporal states: the kicker's run-up and the instant of the kick.
The study encompasses a performance evaluation for eleven distinct HAR backbones, shedding light on their effectiveness in BoGP estimation during free-kick situations. An extra tabular metadata input is introduced, leading to an interesting model enhancement without introducing bias. The promising results reveal 69.1\% accuracy when considering two primary BoGP classes: \textit{right} and \textit{left}. This underscores the model's proficiency in predicting the ball's destination towards the goal with high accuracy, offering promising implications for understanding free-kick dynamics in soccer.}

\onecolumn \maketitle \normalsize \setcounter{footnote}{0} \vfill


\begin{figure*}[t]  
    \centering
    \includegraphics[scale=0.54]{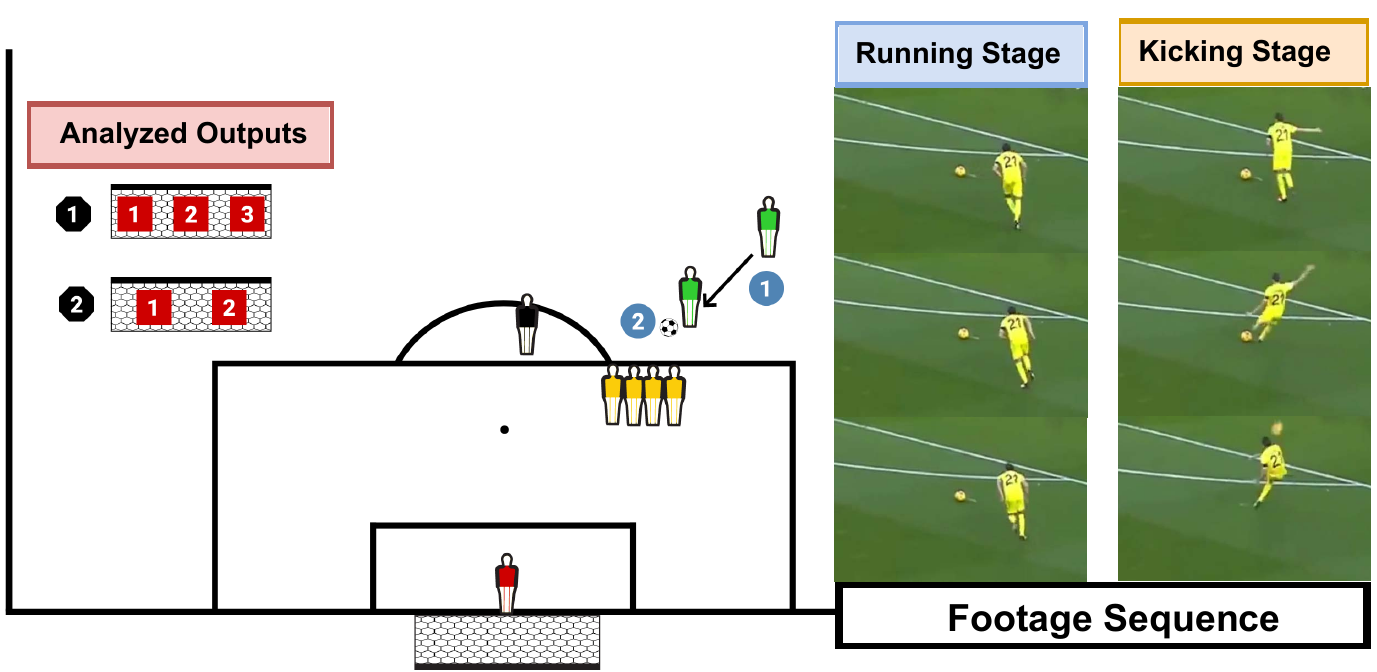}
    \caption{\textbf{Free-kick BoGP classification.} Our proposal involves a thorough analysis of free-kick actions by integrating data from various sources, including free-kick metadata and HAR embeddings. Critically, our classifier combines contextual information with the two-stream action recognition embeddings to make accurate predictions regarding the ball's placement concerning the goal. It is important to note that these experiments relied solely on visual observations of the kicker during the shot without factoring in any ball trajectory data.}
    \label{fig:intro_img}
\end{figure*}

\section{Introduction}
\label{sec:introduction}

In the 2021/22 season, the top 20 revenue-generating clubs collectively made a profit of €9.2 billion, marking a 13\% increase from the previous season and nearly reaching the pre-pandemic levels of 2018/19. This resurgence was driven by the return of fans to stadiums, resulting in a significant increase in matchday revenue, which rose from €111 million to €1.4 billion. The revenue composition of clubs in 2021/22 returned to pre-pandemic levels, with 15\% from matchday activities, 44\% from broadcasting, and 41\% from commercial sources \cite{deloitte2023}. Furthermore, the data indicates that the 2022 FIFA World Cup, held in Qatar, garnered the highest viewership in the tournament's history, with over five billion spectators tuning in through diverse platforms, surpassing more than half of the global population \cite{fifa2022}.

This remarkable financial, as well as the widespread global viewership of soccer events, underscore the tremendous potential and impact of soccer as a mass sport. Furthermore, the evolution of soccer continues after these outstanding statistics. The introduction of technology into the sport is emerging as a pivotal factor, shaping both its on-field dynamics and off-field engagement. According to Microsoft, during a match, players navigate the entire field at high speed, necessitating the deployment of up to 16 fixed cameras for optical tracking positioned around the perimeter of each stadium, capturing a staggering 3.5 million data points per game \cite{microsoftlaliga}. This data is subsequently processed through the Mediacoach platform, making it accessible to clubs and fans through match broadcasts and digital content. Microsoft also remarks that the data strategy is designed to give clubs invaluable insights for adapting training schedules, scrutinizing opponents, and preparing for match days. 

In this context, the integration of technology into soccer has brought about a significant transformation in how the sport is played, assessed, and enjoyed. Several studies and technological innovations have highlighted the potential of technology to enhance various aspects of soccer. Notably, some studies introduced a visual analytic system that combines video recordings with abstract visualizations of trajectory data, enabling analysts to delve deep into ball, player, or team behavior \cite{Stein18,Kamble19,He22}. Furthermore, some comprehensive datasets have been introduced to facilitate the localization of crucial events within extended soccer video footage \cite{Giancola18,Deliege20}. In addition, an automatic method was proposed to localize sports fields in broadcast images, eliminating the need for manual annotation or specialized cameras \cite{Homayounfar17}. Lastly, some analytic systems were developed to visually represent the spatiotemporal evolution of team formations, aiding analysts in understanding and tracking the dynamic aspects of soccer strategies \cite{Wu19,Li23}. These technological advancements have notably transformed sports analysis and enhanced the fan experience in soccer, revealing new insights and engagement opportunities. Nevertheless, unexplored possibilities persist. While previous studies have enriched our understanding of the game, untapped areas exist where technology can drive substantial advancements in soccer. For instance, incorporating predictive analytics in free-kick actions could lead to the creation of advanced algorithms that account for factors like goal distance, angle, kicker skills, defensive wall positioning, and even the goalkeeper's historical performance in stopping free kicks.

This work represents a significant step in advancing our understanding of ball-on-goal position (BoGP) in the context of free kicks directed toward the opponent's goal. Utilizing HAR backbones, we have crafted a BoGP classifier, benchmarking our models against a novel and extensive collection of free-kicks. To accomplish this, we have gathered and processed free-kick footage from various sources on the Internet. Building upon this dataset, multiple models that integrated contextual information and utilized pre-trained HAR encoders (commonly referred to as backbones) were tested to predict the final destination of the kicked ball into the goal. Notably, our methodology incorporates two crucial stages as inputs to the model: the running and the kicking stages, both depicted in Figure \ref{fig:intro_img}.

The significance of this approach lies in the fact that it captures the dynamic nature of a free-kick, allowing our classifier to consider the player's approach and the moment of impact. This nuanced perspective is pivotal for a more accurate and comprehensive understanding of BoGP in free kicks.
Furthermore, we conducted two distinct analyses. The first analysis involved categorizing the goal into three classes (\textit{left}, \textit{center}, and \textit{right}), providing a fine-grained BoGP assessment. The second analysis simplified the categorization into two classes (\textit{left} and \textit{right}), allowing for a broader perspective on BoGP accuracy. This dual approach enabled a deeper exploration of free-kick complexities; please refer to Figure \ref{fig:intro_img}.

Our contributions can be summarized as follows:

\begin{itemize}
    \item We introduce a novel soccer free-kick dataset comprising 603 short clips from actual matches. This dataset has been curated from online sources and is readily accessible to the public.
    \item Through a series of experiments, we empirically showcase the feasibility of addressing the BoGP challenge by employing a classifier that combines contextual data with a two-stream approach. Each stream offers a distinct embedding path, encompassing the running stage and the kicking stage of the free-kick process.
    \item Within the scope of this study, we conduct a comparative analysis of eleven different HAR backbone architectures, assessing their respective performance in BoGP classification. 
    \item An in-depth error analysis study was undertaken to evaluate how the various classes influence the performance of the top-performing model.
\end{itemize}

The subsequent sections of this paper are structured as follows. Section~\ref{sec:relatedwork} discusses previous related work. Section~\ref{sec:pipeline} outlines the proposed pipeline. Section~\ref{sec:results} details the experimental setup and presents the results. Section~\ref{sec:errana} offers an analysis of errors. Lastly, Section~\ref{sec:con} draws our conclusions.

\section{Related work}
\label{sec:relatedwork}


Sports analysis has consistently captured the community's attention, leading to a substantial surge in published research over the past decade. In this sporting domain, technology has become an integral and transformative force, significantly shaping our understanding of sports, as well as how athletes train and compete. This section offers a comprehensive examination of two specific elements addressed in this study: datasets in sports and their computing application.

The available sports video datasets can be categorized into two main groups: still-image and video-sequence datasets.
The first group encompasses datasets primarily designed for image classification. For instance, the UIUC Sports Event Dataset comprises 1,579 images spanning eight sports event categories \cite{Li07}. Each category may contain subsets of images ranging from 180 to 205, categorized as easy or medium based on human subject judgments. Another noteworthy collection is the Leeds Sports Pose Dataset \cite{Johnson10}, featuring 2.000 pose-annotated images of athletes gathered from the Internet. Each image includes annotations for 14 joint locations. More recently, ultra-distance runners competitions have also been captured in wild conditions~\cite{Penate20-prl}.

In contrast, the video-sequence datasets offer time series information about the actions occurring within the scene. These sequences are typically captured using stationary cameras. Sequences from individual sports provide a suitable context for activity recognition, while sequences from team sports can be used for player tracking and event detection. In this context, many sports datasets have been assembled from international competitions to advance research in automatic quality assessment for sports. Some of the most recent datasets include the MTL-AQA diving dataset \cite{Parmar19}, the UNLV AQA-7 dataset, which includes diving, gymnastic vaulting, skiing, snowboarding, and trampoline \cite{Parmar19wacv}, and the Fis-V skating dataset \cite{Xu20}. These datasets have been collected in controlled, non-obstructed environments, with exceptions like the UNLV AQA-7 snowboarding and skiing subsets, gathered in quiet conditions with a dark sky (night) and snowy ground.

The semantic structure of sports video content can be categorized into four layers: raw video, object, event, and semantic layers \cite{Shih18}. The foundation of this pyramid consists of raw video input, from which objects are identified in the higher layers. Specifically, critical objects featured in video clips are recognized through object extraction, such as players \cite{Tianxiao20} and object tracking, including the ball \cite{Shaobo19} and players \cite{JungSoo20}. The event layer signifies the actions of critical objects. Various actions, combined with scene information, generate event labels that depict the related actions and interactions among multiple objects. Research in areas like action recognition \cite{freire22icpr}, re-identification \cite{Akan23,FreireIJCB23}, facial expression recognition \cite{Brick18,ojsantana22mtool}, trajectory prediction \cite{Teranishi20}, and highlight detection \cite{Gao20} falls within the scope of this layer. The topmost layer, the semantic layer, is responsible for summarizing the semantic content of the footage \cite{Cioppa18}. As our objective is BoGP, we seek to classify the outcome of a free-kick action. Furthermore, the mentioned collections predominantly feature professional athletes. In this context, our work does not address the team dimension, as it specifically focuses on a particular action. Nevertheless, several pivotal individuals are visible during this action, including the kicker, the referee, the other players, especially those forming the defensive wall, and the goalkeeper.

\begin{figure*}[t]  
    \centering
    \includegraphics[scale=0.80]{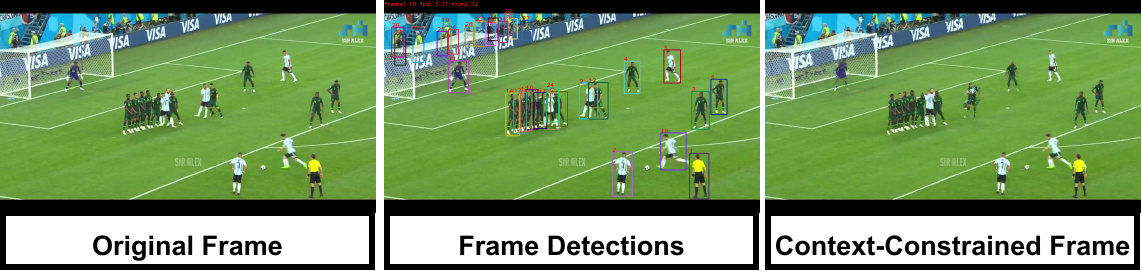}
    \caption{\textbf{Context removal.} For every frame at time $t$, the process entails isolating the kicker's bounding box, which is then superimposed onto a stable background derived from the mean of $\tau$ frames.}
     \label{fig:contextcons}
\end{figure*}

\begin{figure}[t]  
\begin{minipage}[t]{0.9\linewidth}
    \centering
    \includegraphics[scale=0.53]{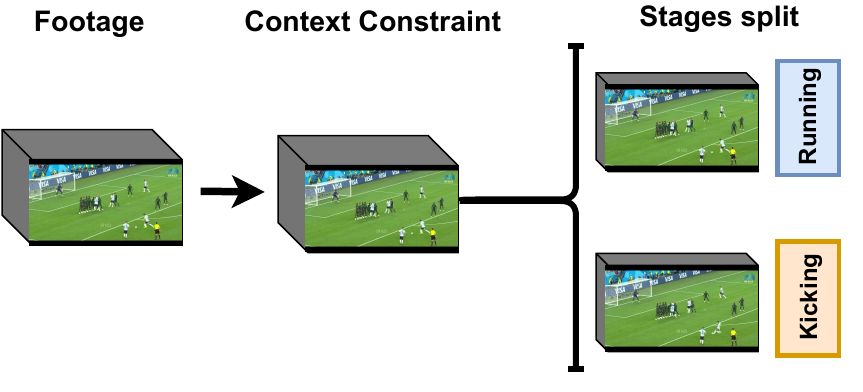}
      \end{minipage}
    \caption{\textbf{Video pre-processing module.} The initial video material undergoes a pre-processing phase wherein the kicker is separated from a dynamic background. Following this, two sets of frames are manually chosen to delineate the running and kicking stages. The remaining frames are excluded.}
     \label{fig:footageprepro}
\end{figure}

\begin{figure*}[t]  
    \centering
    \includegraphics[scale=0.60]{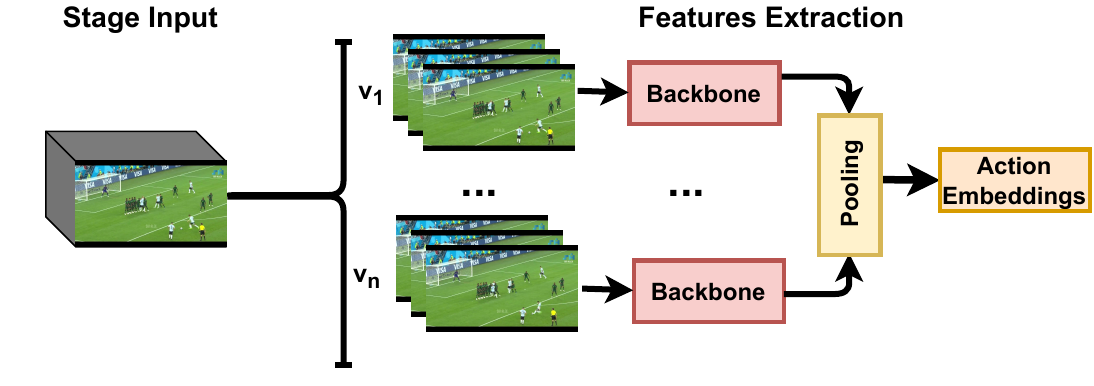}
    \caption{\textbf{Embeddings extraction module.} Each stage footage undergoes downsampling, dividing it into $n$ smaller clips. A pre-trained human-action model is then applied to extract features from these clips. These features are combined using a pooling technique, resulting in a final tensor that serves as input to the classifier. This work examines two pooling methods, average pooling and max pooling.}
     \label{fig:embeddings}
\end{figure*}

\section{Description of the proposal}
\label{sec:pipeline}
This paper introduces and assesses a sequential pipeline consisting of two core modules, where video pre-processing is performed manually before entering the pipeline. The core modules include a video pre-processing module, a stage-embeddings extraction module, and a classifier. Figures \ref{fig:embeddings}, and \ref{fig:classifier} depict visual representations of these modules, while Figure  \ref{fig:footageprepro} illustrates the executed video pre-processing. The following subsections comprehensively describe the video pre-processing step and each module.

\subsection{Context Constraint}
\label{sec:contextconstraint}
In order to optimize the quality of the embeddings generated by the backbone, it is imperative to ensure that the input footage provided to the action recognition networks is devoid of extraneous elements, as indicated in a prior study~\cite{freire22icpr}. Within the context of the dataset utilized for the experiments detailed in this research, as described in Section \ref{sec:dataset}, these extraneous elements encompass unrelated players, staff, supporters, and referees. Given their lack of relevance within the purpose of this work, an initial pre-processing phase is conducted to refine the raw input data by isolating the primary subject, i.e., the kicker. This task is accomplished by leveraging ByteTrack~\cite{zhang2021bytetrack}, a multi-object tracking network that can precisely track the kicker within each video footage, see Figure \ref{fig:contextcons}. Following this, a context-constrained pre-processing technique is applied to establish an ideal setting for conducting the experiments.

In the context of acquiring context-constrained video frames for a specific kicker ($k$) at a given time ($t$) within a specified time interval ($[0, T]$), the bounding box ($BB_k(t)$) plays a crucial role. This bounding box outlines the area occupied by kicker $k$ within the frame recorded at time $t$. To facilitate this process, two primary factors are considered: the bounding box area of the kicker ($BB_k(t)$) and the average number of frames required ($\tau$) to establish a static background against which the isolated kicker ($k$) appears in the pre-processed video frame. The resulting pre-processed frame ($F'_k(t)$) is generated through the following equation:
\[F'_k = BB_{k}(t) \cup \tau\]
Here, the $\cup$ operation involves aligning and superimposing the bounding box of kicker $k$ onto the average of the selected $\tau$ frames. This sequence of pre-processed frames constitutes the new video footage, with the kicker as the sole moving element. 

Lastly, as depicted in Figure \ref{fig:footageprepro}, the resultant footage is temporally segmented. This manual segmentation identifies two distinct moments aligned with the kicker's actions: the running stage and the kicking stage. Any elements in the video, such as the free-kick outcome or the kicker's reaction, have been excluded from the analyzed stages. This study focuses exclusively on the running stage (the phase in which the kicker approaches the ball) and the kicking stage (comprising the 16 frames before and the 16 frames after the ball is kicked).

\subsection{Stage-Embeddings extraction}
\label{sec:embedd}

The preprocessed input footage for each stage, consisting of $m$ frames, undergoes a twofold procedure. Initially, the footage is downsampled, which results in its division into $n$ video clips, represented as ${v_1, ..., v_n}$, where each clip comprises a sequence of $q$ consecutive frames that encapsulate a snapshot of the activity, see Figure \ref{fig:embeddings}. In practical terms, the $n$ clips exhibit partial overlap, spaced one frame apart from the preceding one. These video clips traverse a pre-trained HAR encoder (backbone), producing r-dimensional feature vectors. It is worth noting that these encoder models have undergone prior training on the Kinetics 400 dataset, which encompasses a broad spectrum of 400 action categories~\cite{Kay17}. Following the acquisition of feature vectors for all $n$ video clips, a pooling layer ensures the contribution from each clip. In this regard, we have evaluated both average and max pooling layers, as seen in Section \ref{sec:results}.

\begin{figure*}[t]  
    \centering
    \includegraphics[scale=0.60]{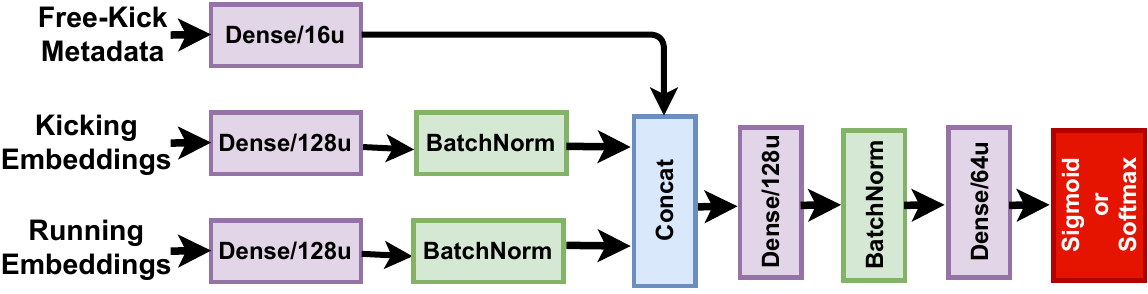}
    \caption{\textbf{The proposed classifier.} 
    Features from the HAR backbones for running and kicking stages are processed alongside free-kick metadata, combining information from various sources to contribute to the model's decision-making process. The features extracted from the HAR backbone offer a fine-grained understanding of the kicker's movements. At the same time, free-kick metadata provides valuable context, influencing the classification outcome, particularly in diverse free-kick scenarios.}
     \label{fig:classifier}
\end{figure*}

We have chosen eleven backbones to test our approach to tackle the BoGP problem. Some are more complex backbones (Slowfast or I3D) than others (the X3D instances and C2D). This section offers an overview of the HAR models considered for this study. The \textbf{C2D} (Convolutional 2D) model, designed for video action classification \cite{C2D14}, exploits the power of 2D Convolutional Neural Networks (CNN) for spatial feature extraction from video frames. Its architecture comprises convolutional layers, pooling layers, and fully connected layers. Convolutional layers extract spatial features while pooling layers reduce dimensionality to prevent overfitting. The C2D model processes each frame independently, employing CNNs to extract spatial features, which are combined to capture temporal action dynamics.

In contrast to the C2D model, the \textbf{SlowFast} model is conceived based on the principle that different video segments possess diverse temporal resolutions and contain crucial information for action recognition \cite{SlowFast19}. For example, some actions occur swiftly and necessitate high temporal resolution for detection, while others unfold more slowly and can be recognized with a lower temporal resolution. To address this variability, the SlowFast model adopts a dual-pathway approach, comprising fast and slow pathways that operate on video data at varying temporal resolutions.

Similarly, \textbf{Slow} adopts a two-stream architecture to capture both short-term and long-term temporal dynamics in videos~\cite{Slow21}. Its slow pathway processes high-resolution frames but at a lower frame rate, similar to the C2D model. Additionally, Slow incorporates a temporal-downsampling layer to capture longer-term temporal dynamics. The Inflated 3D ConvNet (I3D) model is designed to handle short video clips as 3D spatiotemporal volumes, enabling the capture of both appearance and motion cues using a two-stream approach~\cite{Carreira17}. In this design, the first stream deals with RGB images, utilizing weights that are pre-trained on extensive image classification datasets. Simultaneously, the second stream processes optical flow images and undergoes fine-tuning in conjunction with the RGB stream.

A revised variant of the I3D model, \textbf{I3D NLN}, incorporates non-local operations to enhance spatiotemporal dependency modeling in videos~\cite{NonlocalNN17}. I3D NLN retains the two-stream architecture involving RGB and optical flow streams, processing 3D spatiotemporal volumes. In contrast to the Inception module, I3D NLN employs non-local blocks capable of learning long-range dependencies across feature map positions. By computing weighted sums of input features from all positions based on the similarity between these positions in the feature maps, I3D NLN captures global context information and improves the modeling of temporal dynamics.

Finally, we have leveraged four \textbf{X3D} model variations, distinguished by their sizes: extra small (X3D-XS), small (X3D-S), medium (X3D-M), and large (X3D-L). Each expansion incrementally transforms X2D from a compact spatial network to a spatiotemporal X3D network \cite{Feichtenhofer20} by modifying temporal (frame rate and sampling rate), spatial (footage resolution), width (network depth), and depth dimensions (number of layers and units). X3D-XS results from five expansion steps, followed by X3D-S, which includes one backward contraction step after the seventh expansion. X3D-M and X3D-L are generated by the eighth and tenth expansions, respectively. X3D-M augments the spatial resolution by elevating the spatial sampling resolution of the input video. At the same time, X3D-L expands the spatial resolution and network depth by increasing the number of layers in each residual stage.

\begin{figure*}[t]  
    \centering
    \includegraphics[scale=0.75]{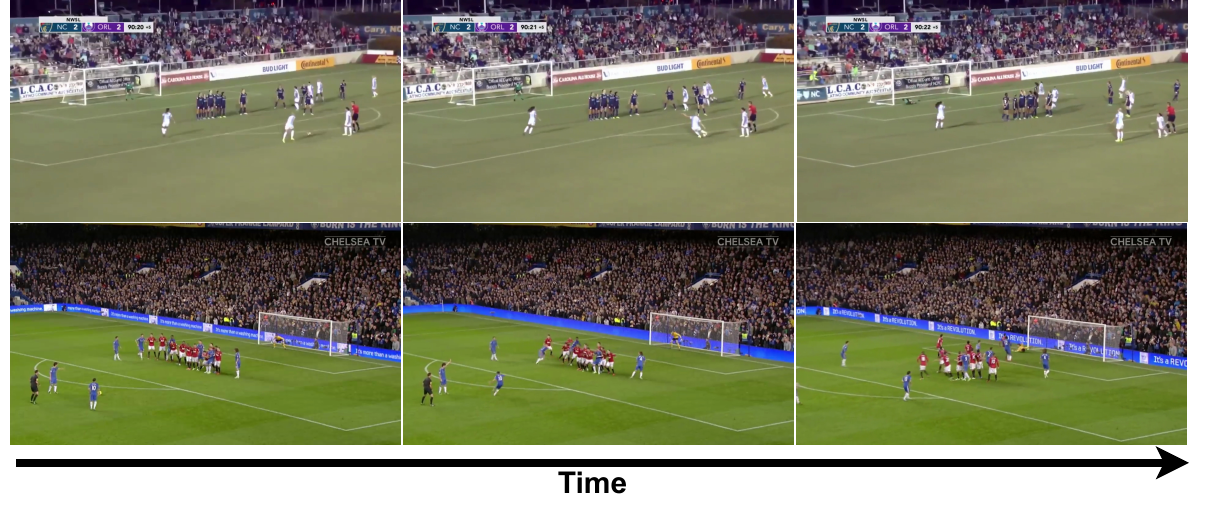}
    \caption{\textbf{Free-kick dataset sequences.} The video dataset used in this study was gathered from the Internet without any imposed usage restrictions. Due to this unrestricted collection approach, the dataset exhibits notable pose, scale, and lighting conditions variability. Video clips were carefully edited to retain frames from just before the running stage until the moment of the free-kick outcome.}
     \label{fig:dataset}
\end{figure*}

\subsection{Classifier}
\label{sec:classifier}
The proposed classifier involves feature extraction from the identical HAR backbone for both the running and kicking stages, as well as the inclusion of free-kick metadata, see Figure \ref{fig:classifier}. 

This three-input approach combines information from various sources, each contributing unique and complementary insights to the model's decision-making process. The features extracted from the HAR backbone offer a fine-grained understanding of the kicker's movements and actions during the free kick. Simultaneously, free-kick metadata provides valuable context and situational information that can significantly influence the classification outcome, especially when dealing with various free-kick scenarios. In this regard, the free-kick metadata encompasses four distinct input variables, each contributing specific information to the model's decision-making process. These variables include pitch side, free-kick side, free-kick distance, and kicker foot. The pitch side variable operates as a binary indicator, distinguishing between left and right. In contrast, the free-kick side variable offers a more detailed classification, representing three distinct values related to the shooting point: left to the goal, center to the goal, and right to the goal. Similarly, free-kick distance, another binary variable, provides insight into whether the free kick occurs near or far from the penalty box. Lastly, the kicker foot variable, also binary, characterizes the preferred kicking foot as either left or right.

As a result, the model receives three distinct inputs, each of which undergoes processing via dedicated fully connected layers with varying units (16 and 128) based on the nature of the input. The running and kicking paths also include batch normalization layers. Subsequently, all paths are concatenated, followed by two fully connected layers (128 and 64 units, respectively), separated by a batch normalization layer. Finally, the model's output, denoting the ball's position on the goal, is determined by either a Sigmoid or a Softmax layer, depending on whether the output comprises two or three classes.

In the conventional classification framework, the primary objective is to assign a sample to its appropriate class. In this context, we have conducted two experiments on the ball's positioning within the goal. The first experiment considers three distinct classes (\textit{left}, \textit{right}, and \textit{center}), while the second experiment operates as a binary classifier, explicitly distinguishing between \textit{left} and \textit{right} placements. Consequently, we employ the categorical cross-entropy loss function for the first experiment:

\begin{equation}
    Loss_{1} = -\sum_{i=1}^{C} y_i \log(p_i)
\end{equation}

Where $C$ is the number of classes, $y_i$ is the true probability distribution (one-hot encoded vector) of the ground truth class, and $p_i$ is the predicted probability for class $i$.
For the second experiment, the considered loss function to tackle the problem is the binary cross-entropy:

\begin{equation}
    Loss_{2} = \frac{-1}{N}\sum_{i=1}^{N}-{(y_i\log(p_i) + (1 - y_i)\log(1 - p_i))}
\end{equation}

Where $p_i$ is the i-th scalar value in the model output, $y_i$ is the corresponding target value, and N is the number of scalar values in the model output.

\section{Experimental setup}
\label{sec:results}
This section is divided into three subsections related to the dataset acquisition, experimental setup, and achieved results of the designed experiments. The first subsection provides technical details regarding the dataset, including its acquisition and data-cleaning processes. The second subsection outlines the technical aspects of our proposal, such as the data split. Finally, the third subsection summarizes the achieved results.

\subsection{Dataset}
\label{sec:dataset}
To our knowledge, there is no publicly available soccer free-kick dataset. Our data collection approach hinges on generality, intending to construct robust detection models for practical use. This compilation of videos was sourced from the Internet without any usage restrictions, resulting in considerable variations in pose, scale, and lighting conditions, see Figure~\ref{fig:dataset}. The data collection process encompasses three steps:

\begin{enumerate}
    \item \textbf{Web Scraping:} an extensive search was conducted to gather relevant images using keywords like "free-kick soccer," "free-kick compilation", and the names of various soccer players well known for frequently shooting free kicks.
    
    \item \textbf{Shot Labeling:} labeling involves carefully editing each video clip. These clips are trimmed to cover the period from just before the kicker initiates the run to the occurrence of the shot outcome. This stage results in a subset of 603 free-kick clips.
    
    \item \textbf{Manual Annotation:} each free-kick clip is manually reviewed and annotated. Annotations encompass various variables, including pitch side, free-kick side, free-kick distance, kicker foot (left or right), kick outcome, barrier configuration, gender, goalkeeper zone, and the specific frame in which the ball is kicked. The resolution of these clips is $1920\times1080$ pixels.
\end{enumerate}

\begin{table*}[!htbp]
 \centering
 \begin{tabular}{|l|c|c|c|c|c|c|}
 \hline
  \small Backbone & \small \#Frames &\small Pooling &\small Accuracy & \small Precision & \small Recall & \small F1-Score\\
 \hline
 \hline 
 \footnotesize C2D~\cite{C2D14} & \footnotesize 8  & \footnotesize Average &\footnotesize 52.9\% &\footnotesize 49.4\% &\footnotesize 43.1\% &\footnotesize 46.1\% \\ 
 \hline
 
 I3D~\cite{Carreira17} & \footnotesize 8  &\footnotesize Average & \footnotesize 51.4\% & \footnotesize 42.7\% & \footnotesize 39.6\% & \footnotesize 41.1\% \\
 \hline
 
 I3D NLN~\cite{NonlocalNN17} & \footnotesize 8  &\footnotesize Average & \footnotesize  51.9\% & \footnotesize  44.6\% & \footnotesize  41.2\% & \footnotesize  42.8\% \\
 \hline

Slow4x16~\cite{Slow21} &\footnotesize 4  &\footnotesize Average & \footnotesize  55.0\% & \footnotesize  49.4\% & \footnotesize  44.6\% & \footnotesize  46.9\% \\
 \hline
 Slow8x8~\cite{Slow21} &\footnotesize 8  &\footnotesize Average &\footnotesize 55.3\% &\footnotesize 46.1\% &\footnotesize 41.5\% & \footnotesize 43.7\% \\ 
 \hline
 SlowFast4x16~\cite{SlowFast19} &\footnotesize 32  &\footnotesize Max & \footnotesize 55.0\% & \footnotesize 47.1\% & \footnotesize 43.9\% & \footnotesize 45.4\% \\
 \hline
 SlowFast8x8~\cite{SlowFast19} &\footnotesize 32  &\footnotesize Average & \footnotesize  53.4\% & \footnotesize  47.4\% & \footnotesize  45.2\% & \footnotesize  46.2\% \\
 \hline
  X3D-XS~\cite{Feichtenhofer20} &\footnotesize 4 &\footnotesize Max & \footnotesize  51.2\% & \footnotesize  46.3\% & \footnotesize 43.9\% & \footnotesize  45.1\% \\
  \hline
  X3D-S~\cite{Feichtenhofer20} &\footnotesize 4 &\footnotesize Max & \footnotesize  53.4\% & \footnotesize  44.9\% & \footnotesize 43.5\% & \footnotesize  44.2\% \\
  \hline
  X3D-M~\cite{Feichtenhofer20} &\footnotesize 13 &\footnotesize Max & \footnotesize  53.6\% & \footnotesize  47.9\% & \footnotesize 43.0\% & \footnotesize  45.3\% \\
  \hline
  X3D-L~\cite{Feichtenhofer20} &\footnotesize 16 &\footnotesize Average & \textbf{\footnotesize  57.2\%} & \textbf{\footnotesize  50.0\%} & \textbf{\footnotesize 48.5\%} & \textbf{\footnotesize  49.3\%} \\
 \hline
 \end{tabular}
  \caption{\textbf{Comparative performance analysis of HAR architectures for BoGP estimation when considering three classes}. This table compares different backbone architectures used to detect BoGP during free-kick shots. The first column lists the backbone models, while the second column specifies the number of frames the model utilizes for generating HAR embeddings. The table includes crucial performance metrics such as the number of frames per embedding backbone, the applied pooling method, and the values of the performance metrics: accuracy, precision, recall, and F1-Score.
 }
  \label{tab:3cls_shoot_zone}
 \end{table*}

Despite the initial inclusion of 603 free-kick clips in the dataset, several factors reduced this number. A critical consideration was the camera viewpoint, as it played a substantial role in the selection process. To maintain shooting action stability, clips where the camera perspective was positioned behind the goalkeeper or the kicker were excluded. As described in Section \ref{sec:contextconstraint}, the remaining 584 videos underwent people detection using ByteTrack. Unfortunately, some videos exhibited low image quality, resulting in subpar detection performance. As a consequence, the dataset was further reduced to 539 clips.

Subsequently, the duration of the videos became a focal point, as clips that were excessively short in length were found to be inadequate for extracting meaningful information. For instance, videos commencing precisely as the player initiated the kick (without a preceding running stage) were omitted from consideration due to the need for a minimum frame count to extract pertinent information. All clips containing fewer than 32 frames were accordingly excluded, ultimately reducing the dataset to 451 clips.

The problem's intrinsic nature also emerged as a significant determining factor during clip selection. Specifically, any clips in which the kick did not successfully reach the goal, such as instances where the ball failed to surpass the defensive barrier, were omitted. In such cases, it was infeasible to ascertain the target location within the goal, rendering these clips inapplicable. Therefore, a refined subset of 418 clips was designated for inclusion in this study.

\subsection{Experimental setup}

The results presented in this section refer to the average accuracy on five repetitions of 10-fold cross-validation for each experiment. Significantly, the class distribution within the dataset is characterized as follows: 187 free-kick shots are directed towards the left side of the goal, 181 are aimed at the right side, and 50 target the center area of the goal. The class distribution exhibits a notable imbalance, particularly in the case of the center-side shots. We have implemented a class weighting strategy during the model training phase to address this issue. The adjustment of class weights in the training process serves to amplify the model's sensitivity to minority classes, effectively mitigating the inherent challenge of disparate class distributions. This approach serves as a valuable mechanism to rectify any potential bias arising from the overrepresentation of majority classes, thereby ensuring equitable model performance across all classes.

\begin{table*}[!htbp]
 \centering
 \begin{tabular}{|l|c|c|c|c|c|c|}
 \hline
  \small Backbone & \small \#Frames &\small Pooling &\small Accuracy & \small Precision & \small Recall & \small F1-Score\\
 \hline
 \hline 
 \footnotesize C2D~\cite{C2D14} & \footnotesize 8  & \footnotesize Max &\footnotesize 67.4\% &\footnotesize 56.5\% &\footnotesize 60.2\% &\footnotesize 58.3\% \\ 
 \hline
 
 I3D~\cite{Carreira17} & \footnotesize 8  &\footnotesize Average & \footnotesize 63.1\% & \footnotesize 51.3\% & \footnotesize 56.4\% & \footnotesize 53.7\% \\
 \hline
 
 I3D NLN~\cite{NonlocalNN17} & \footnotesize 8  &\footnotesize Max & \footnotesize  62.8\% & \footnotesize  52.6\% & \footnotesize  51.7\% & \footnotesize  52.2\% \\
 \hline
 
Slow4x16~\cite{Slow21} &\footnotesize 4  &\footnotesize Average & \footnotesize  66.9\% & \textbf{\footnotesize  60.0\%} & \footnotesize  68.3\% & \footnotesize  63.9\% \\
 \hline
 
 Slow8x8~\cite{Slow21} &\footnotesize 8  &\footnotesize Max &\footnotesize 65.8\% &\footnotesize 57.6\% &\footnotesize 67.7\% & \footnotesize 62.2\% \\ 
 \hline
 SlowFast4x16~\cite{SlowFast19} &\footnotesize 32  &\footnotesize Average & \textbf{\footnotesize 69.1\%} & \footnotesize 57.7\% & \textbf{\footnotesize 76.1\%} & \textbf{\footnotesize 65.7\%} \\
 \hline
 SlowFast8x8~\cite{SlowFast19} &\footnotesize 32  &\footnotesize Max & \footnotesize  63.6\% & \footnotesize  56.6\% & \footnotesize  69.9\% & \footnotesize  62.5\% \\
 \hline
  X3D-XS~\cite{Feichtenhofer20} &\footnotesize 4 &\footnotesize Max & \footnotesize  61.9\% & \footnotesize  47.2\% & \footnotesize 48.3\% & \footnotesize  47.7\% \\
  \hline
  X3D-S~\cite{Feichtenhofer20} &\footnotesize 4 &\footnotesize Average & \footnotesize  64.4\% & \footnotesize  50.9\% & \footnotesize 74.3\% & \footnotesize  60.4\% \\
  \hline
  X3D-M~\cite{Feichtenhofer20} &\footnotesize 13 &\footnotesize Max & \footnotesize  66.0\% & \footnotesize  58.4\% & \footnotesize 51.4\% & \footnotesize  54.7\% \\
  \hline
  X3D-L~\cite{Feichtenhofer20} &\footnotesize 16 &\footnotesize Max & \footnotesize  65.8\% & \footnotesize  59.7\% & \footnotesize 56.6\% & \footnotesize  58.1\% \\
 \hline
 \end{tabular}
  \caption{\textbf{Comparative performance analysis of HAR architectures for soccer player free-kick shoot zone estimation when considering two classes}. This table compares different backbone architectures used to detect soccer player shoot zones during free-kick shots. The first column lists the backbone models, while the second column specifies the number of frames the model utilizes for generating HAR embeddings. The table includes crucial performance metrics such as the number of frames utilized, the pooling method applied, accuracy, precision, recall, and F1-Score. These metrics offer valuable insights into the effectiveness of each backbone architecture for this specific task.
 }
  \label{tab:2cls_shoot_zone}
 \end{table*}

\subsection{Results}

Table \ref{tab:3cls_shoot_zone} presents a comparative performance analysis of various HAR backbone architectures utilized to estimate the BoGP during free-kick shots, specifically when considering three different target classes: \textit{left}, \textit{center}, and \textit{right}. The table highlights the number of frames used for each embedding backbone (denoted as $q$ in Section \ref{sec:embedd}), the pooling method employed, and key performance metrics including accuracy, precision, recall, and F1-Score.
The presented HAR backbone architectures encompass a range of models described in Section \ref{sec:embedd}. Each model is evaluated based on the aforementioned metrics, providing valuable insights into their effectiveness in BoGP estimation during free-kick situations.

A noteworthy observation pertains to the choice of pooling layers for the HAR embeddings (see Figure \ref{fig:embeddings}). The data presented in Table \ref{tab:3cls_shoot_zone} reveals an intriguing trend: lighter models, exemplified by the X3D instances, tend to favor the utilization of the MaxPool layer, while heavier models typically demonstrate a preference for the AveragePool layer. This distinction in pooling layer selection reflects these models' diverse architectural considerations and requirements, underscoring the need to suit the pooling method to the specific characteristics and demands of a given HAR model.

The table prominently illustrates the distinct performance levels exhibited by various models. X3D-L, in particular, stands out as the top performer, boasting the highest accuracy (57.2\%), precision (50.0\%), recall (48.5\%), and F1-Score (49.3\%). Following closely in classification performance are the SlowFast and Slow instances, although they lag by a margin of 2.2\% in accuracy. It is worth noting that the overall performance in the context of three-class classification remains relatively modest, as evidenced by the F1-Score, though exceeding that of a random classifier. Section \ref{sec:errana} provides a comprehensive error analysis.

To complete our evaluation, Table \ref{tab:2cls_shoot_zone} presents a comparative performance analysis of various HAR backbone architectures used in soccer player free-kick shoot zone estimation when considering two classes: \textit{left} and \textit{right}. Once again, the models are evaluated in this scenario based on details about the number of frames used, the pooling method applied, and the four key performance metrics: accuracy, precision, recall, and F1-Score.
Comparing this table with the previously discussed Table \ref{tab:3cls_shoot_zone}, we observe an interesting transition regarding the number of classes considered. The simplification of the classification task has a notable impact on model performance. Despite the reduced complexity of the classification problem, there are variations in the performance of the backbone architectures, indicating that the choice of backbone remains critical.
Performance-wise, several observations can be made. For instance, SlowFast4x16 exhibits the highest accuracy (69.1\%) in this two-class classification scenario, outperforming other models. Additionally, Slow4x16 achieves a remarkable 60.0\% precision, indicating its ability to accurately classify instances. The F1-Score, which combines precision and recall, is also noteworthy, with SlowFast4x16 leading the way with a score of 65.7\%. These metrics provide valuable insights into the effectiveness of the backbone architectures for the specific task of soccer player free-kick shoot zone estimation. In contrast to the outcomes in the three-class scenario, the utilization of MaxPool and AveragePool layers is evenly distributed in this table.

The architecture of the classifier described in Section \ref{sec:classifier} poses an intriguing question: how does the incorporation of free-kick metadata impact performance? Upon calculating the mean accuracy across all scrutinized models, the obtained outcome indicates that without consideration for free-kick metadata, the accuracy diminishes by 3\%, and the F1-Score experiences a 4\% decline. This signifies that metadata enhances contextual information regarding free-kick embeddings, yet it does not introduce bias to the proposed model.

In summary, as shown in this table, the transition from a three-class to a two-class problem emphasizes the consequences of simplifying the classification task. It underscores the performance differences among various HAR backbone architectures and their potential suitability for specific sports action recognition tasks. However, these findings have raised several questions, including the influence of the center class on classification, the distribution of error predictions, and the examination of confusion matrices for the top-performing models. These questions will be addressed in the following section.

\section{Error Analysis}
\label{sec:errana}
\begin{figure*}[!htb]
   \begin{minipage}{0.48\textwidth}
     \centering
     \includegraphics[width=.8\linewidth]{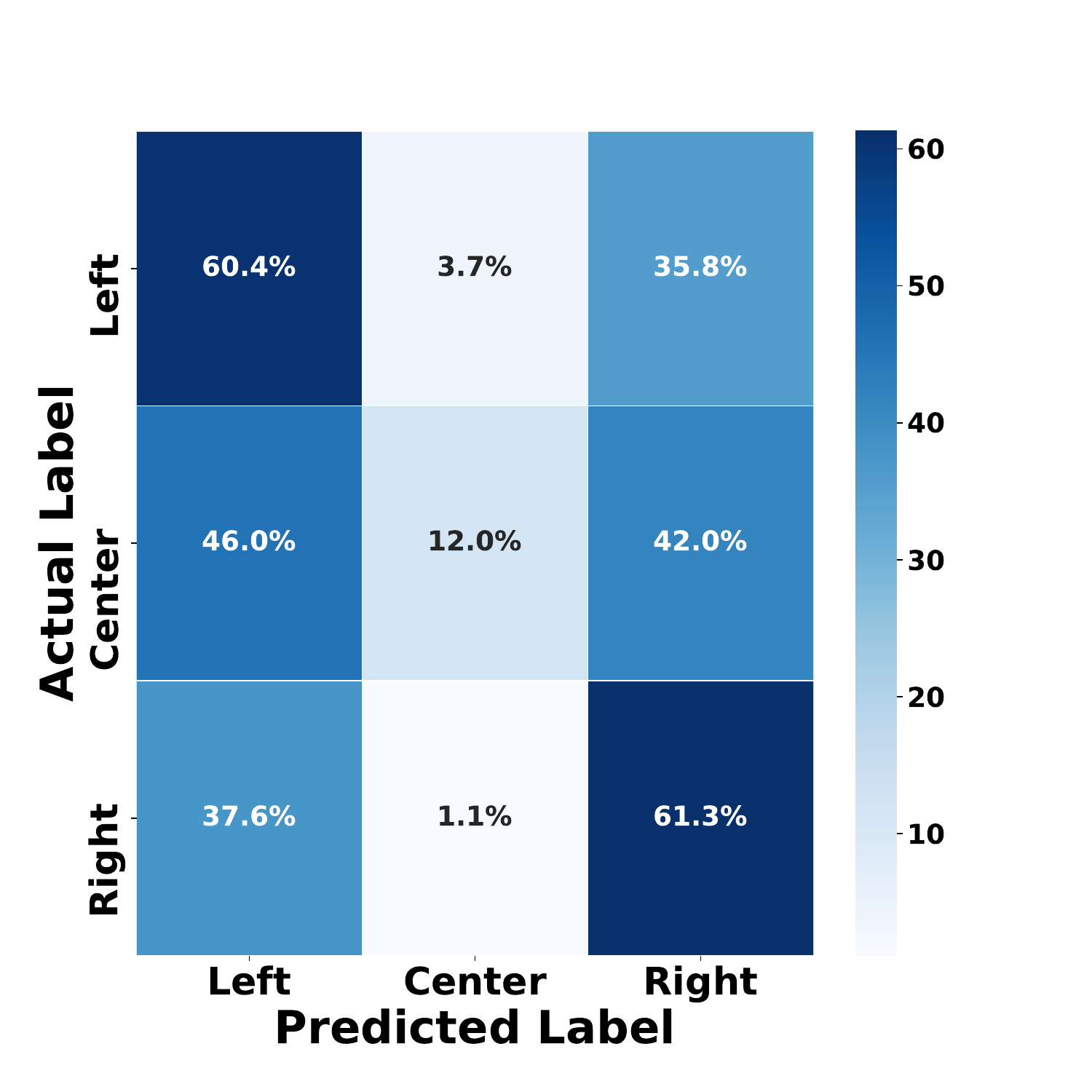}
     \caption{Three-class SlowFast4x16 confusion matrix.}\label{Fig:CF1}
   \end{minipage}\hfill
   \begin{minipage}{0.48\textwidth}
     \centering
     \includegraphics[width=.8\linewidth]{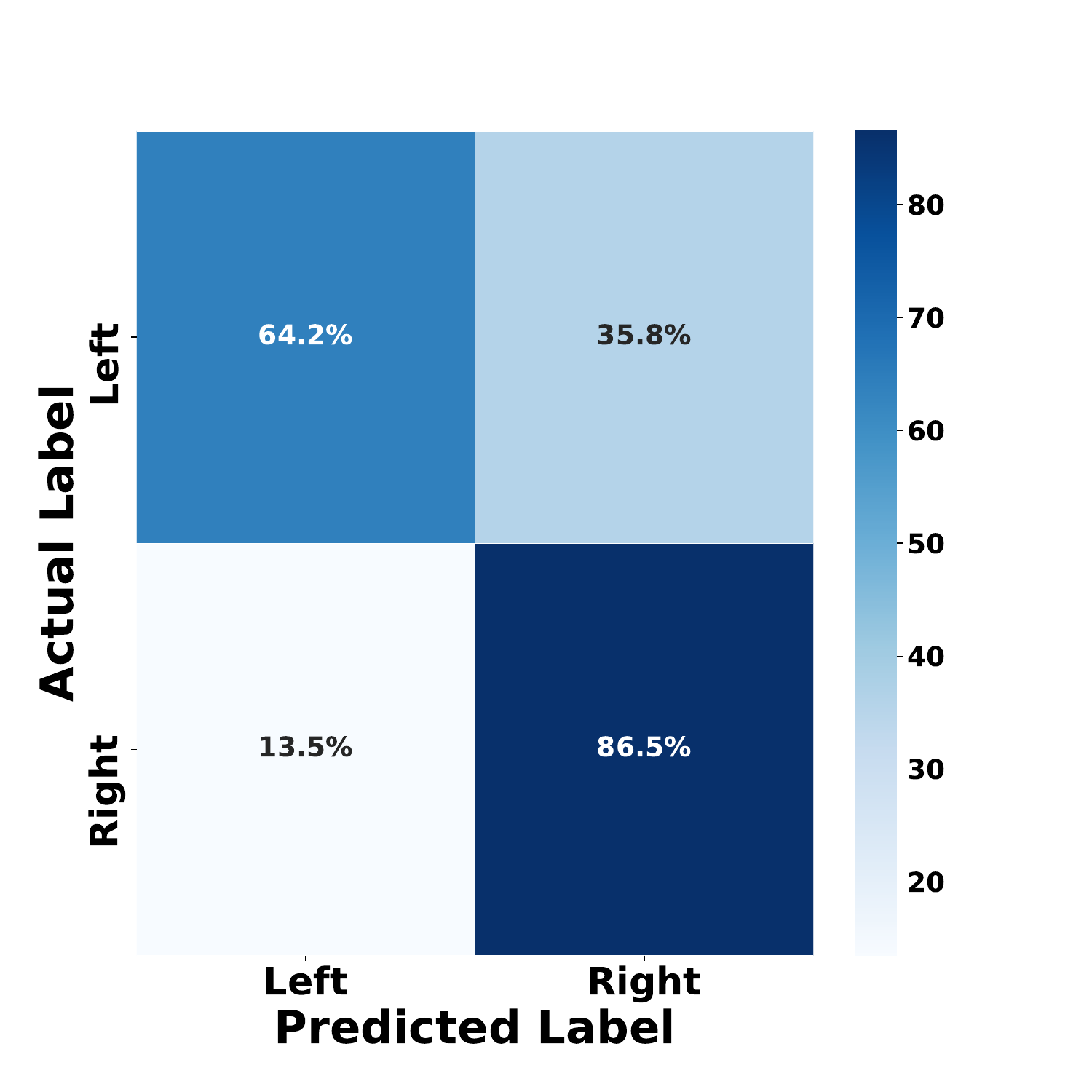}
     \caption{Two-class SlowFast4x16 confusion matrix.}\label{Fig:CF2}
   \end{minipage}
\end{figure*}

In this section, our primary focus is on the top-performing model, which employs the SlowFast4x16 backbone. It is crucial to comprehensively analyze its performance under scenarios involving two and three classes. As a case in point, Figures \ref{Fig:CF1} and \ref{Fig:CF2} visually represent the confusion matrices for both experimental settings.

\begin{figure*}[t]  
    \centering
    \includegraphics[scale=0.75]{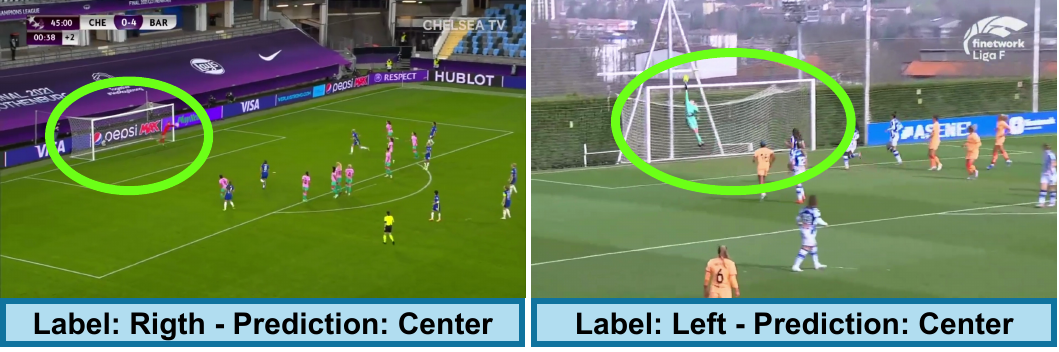}
    \caption{\textbf{SlowFast4x16 misclassified clips.} These frames represent the ultimate phase of two distinct samples. It is important to note that the proposed model exclusively examines the actions of the kicker, meaning it does not consider any frames beyond the 16 post-kicking frames, and the background remains static. Consequently, the frames presented in this figure were never seen by the models; they are included solely to exemplify the intricacies associated with the \textit{center} class. Notably, the classifier erroneously categorizes these clips as \textit{center} when labeled as \textit{right} and \textit{left}, respectively.}
     \label{fig:misclass}
\end{figure*}

Our analysis presents the confusion matrix for our classification model, designed to categorize free-kick soccer actions into one of three classes: \textit{left}, \textit{center}, or \textit{right}. As illustrated in Figure \ref{Fig:CF1}, this matrix provides valuable insight into the model's performance and ability to classify BoGP correctly.
The diagonal elements of the matrix represent instances where the model's predictions align with the actual classes. For instance, the model achieved an accuracy of approximately 60.4\% in identifying \textit{left} shots, 12.0\% for \textit{center}, and 61.3\% for \textit{right}. These values indicate the model's proficiency in correctly classifying shots into their respective categories.
However, the off-diagonal elements reveal cases of misclassification. Notably, there is some confusion between the \textit{center} and the other two classes. The model often misclassifies \textit{center} shots as \textit{left} (46.0\%) or \textit{right} (42.0\%), suggesting improvement in distinguishing \textit{center} shots from the others. Additionally, \textit{left} shots are occasionally misclassified as \textit{right} (35.8\%), and \textit{right} shots are occasionally mislabeled as \textit{left} (37.6\%) or \textit{center} (1.1\%).

Our analysis suggests that the classifier faces challenges in accurately distinguishing the \textit{center} category, as illustrated in Figure \ref{fig:misclass}. The intricacies of this classification become apparent, even for human annotators, as the camera perspective can sometimes obscure the goal's position. This issue is compounded by the limited number of \textit{center} samples, coupled with the wide range of camera angles in the dataset. Consequently, achieving a fine-grained classification for \textit{center} may not be practically feasible given these constraints.

The confusion matrix shown in Figure \ref{Fig:CF2} suggests a notable accuracy in classifying instances, particularly on the diagonal elements. The top-left quadrant indicates a correct classification rate of 64.2\% for the \textit{left} category, while the bottom-right quadrant signifies an 86.5\% accuracy in classifying the \textit{right} category. However, there is some misclassification, as evidenced by the off-diagonal elements, with 35.8\% of \textit{left} instances being erroneously classified as \textit{right} and 13.5\% of \textit{right} instances being misclassified as \textit{left}.

\section{\uppercase{Conclusions}}
\label{sec:con}

In conclusion, this study extensively examined the performance of various HAR backbone architectures in estimating the BoGP during free-kick shots. The investigation covered three-class (\textit{left}, \textit{center}, and \textit{right}) and two-class (\textit{left} and \textit{right}) classification scenarios, providing valuable insights into the effectiveness of different models.

X3D-L emerged as the top performer for the three-class classification with notable accuracy, precision, recall, and F1-Score. However, the overall performance in this context remained modest, prompting a comprehensive error analysis in Section \ref{sec:errana}. In contrast, the two-class scenario revealed a transition in the number of classes and demonstrated that despite the reduced complexity, the choice of backbone architecture remains critical. SlowFast4x16 exhibited the highest accuracy and noteworthy precision and F1-Score, highlighting its effectiveness in soccer player free-kick shoot zone estimation.
The inclusion of Free-kick metadata in the analysis showcased its impact on performance, revealing a 3\% accuracy drop and a 4\% decline in F1-Score when not considered. Importantly, this decline signifies the role of metadata in enhancing contextual information without introducing bias to the model.

The focus on the top-performing model, SlowFast4x16, included a detailed examination of confusion matrices for the three-class and the two-class scenarios. While the model demonstrated proficiency in classifying instances, challenges were identified, particularly in distinguishing the \textit{center} category. The limited number of samples and diverse camera angles posed practical challenges in achieving fine-grained classification for \textit{center}.

These findings highlight the complexity of sports action recognition tasks, emphasizing the need for careful consideration of the model architecture and task simplification's influence. Further questions were raised, including the impact of the center class on classification, the distribution of error predictions, and the exploration of confusion matrices for top-performing models, providing avenues for future research and improvement.

\section*{\uppercase{Acknowledgements}}

This work is partially funded by the Spanish Ministry of Science and Innovation under project PID2021-122402OB-C22 and by the ACIISI-Gobierno de Canarias and European FEDER funds under project ULPGC Facilities Net and Grant \mbox{EIS 2021 04}.


\bibliographystyle{apalike}

{\small
\begin{thebibliography}{}

\bibitem[Akan and Varli, 2023]{Akan23}
Akan, S. and Varli, S. (2023).
\newblock Reidentifying soccer players in broadcast videos using body feature alignment based on pose.
\newblock In {\em Proceedings of the 2023 4th International Conference on Computing, Networks and Internet of Things}, page 440–444, New York, NY, USA. Association for Computing Machinery.

\bibitem[Brick et~al., 2018]{Brick18}
Brick, N.~E., McElhinney, M.~J., and Metcalfe, R.~S. (2018).
\newblock The effects of facial expression and relaxation cues on movement economy, physiological, and perceptual responses during running.
\newblock {\em Psychology of Sport and Exercise}, 34:20--28.

\bibitem[Carreira and Zisserman, 2017]{Carreira17}
Carreira, J. and Zisserman, A. (2017).
\newblock {Quo Vadis, Action Recognition? A New Model and the Kinetics Dataset}.
\newblock In {\em 2017 IEEE Conference on Computer Vision and Pattern Recognition (CVPR)}, pages 4724--4733.

\bibitem[Cioppa et~al., 2018]{Cioppa18}
Cioppa, A., Deliège, A., and Van~Droogenbroeck, M. (2018).
\newblock A bottom-up approach based on semantics for the interpretation of the main camera stream in soccer games.
\newblock In {\em 2018 IEEE/CVF Conference on Computer Vision and Pattern Recognition Workshops (CVPRW)}, pages 1846--184609.

\bibitem[Deli{\`e}ge et~al., 2021]{Deliege20}
Deli{\`e}ge, A., Cioppa, A., Giancola, S., Seikavandi, M.~J., Dueholm, J.~V., Nasrollahi, K., Ghanem, B., Moeslund, T.~B., and Droogenbroeck, M.~V. (2021).
\newblock Soccernet-v2 : A dataset and benchmarks for holistic understanding of broadcast soccer videos.
\newblock In {\em The IEEE Conference on Computer Vision and Pattern Recognition (CVPR) Workshops}.

\bibitem[Deloitte, 2023]{deloitte2023}
Deloitte (2023).
\newblock Deloitte football money league 2023.
\newblock Accessed on November 3, 2023.

\bibitem[Feichtenhofer, 2020]{Feichtenhofer20}
Feichtenhofer, C. (2020).
\newblock {X3D: Expanding Architectures for Efficient Video Recognition}.
\newblock {\em 2020 IEEE/CVF Conf. on Computer Vision and Pattern Recognition (CVPR)}, pages 200--210.

\bibitem[Feichtenhofer et~al., 2018]{SlowFast19}
Feichtenhofer, C., Fan, H., Malik, J., and He, K. (2018).
\newblock Slowfast networks for video recognition.
\newblock {\em 2019 IEEE/CVF International Conference on Computer Vision (ICCV)}, pages 6201--6210.

\bibitem[Feichtenhofer et~al., 2021]{Slow21}
Feichtenhofer, C., Fan, H., Xiong, B., Girshick, R.~B., and He, K. (2021).
\newblock A large-scale study on unsupervised spatiotemporal representation learning.
\newblock {\em 2021 IEEE/CVF Conference on Computer Vision and Pattern Recognition (CVPR)}, pages 3298--3308.

\bibitem[FIFA, 2022]{fifa2022}
FIFA (2022).
\newblock 2019-2022 revenue.
\newblock Accessed on November 3, 2023.

\bibitem[Freire-Obregón et~al., 2022]{freire22icpr}
Freire-Obregón, D., Lorenzo-Navarro, J., Santana, O.~J., Hernández-Sosa, D., and Castrillón-Santana, M. (2022).
\newblock {Towards cumulative race time regression in sports: I3D ConvNet transfer learning in ultra-distance running events}.
\newblock In {\em International Conference on Pattern Recognition (ICPR)}, pages 805--811.

\bibitem[Freire-Obregón et~al., 2023]{FreireIJCB23}
Freire-Obregón, D., Lorenzo-Navarro, J., Santana, O.~J., Hernández-Sosa, D., and Castrillón-Santana, M. (2023).
\newblock {A Large-Scale Re-identification Analysis in Sporting Scenarios: the Betrayal of Reaching a Critical Point}.
\newblock In {\em International Joint Conference on Biometrics (IJCB)}.

\bibitem[Gao et~al., 2020]{Gao20}
Gao, X., Liu, X., Yang, T., Deng, G., Peng, H., Zhang, Q., Li, H., and Liu, J. (2020).
\newblock Automatic key moment extraction and highlights generation based on comprehensive soccer video understanding.
\newblock In {\em 2020 IEEE International Conference on Multimedia Expo Workshops (ICMEW)}, pages 1--6.

\bibitem[Giancola et~al., 2018]{Giancola18}
Giancola, S., Amine, M., Dghaily, T., and Ghanem, B. (2018).
\newblock Soccernet: A scalable dataset for action spotting in soccer videos.
\newblock In {\em 2018 IEEE/CVF Conference on Computer Vision and Pattern Recognition Workshops (CVPRW)}, pages 1792--1810.

\bibitem[Guo et~al., 2020]{Tianxiao20}
Guo, T., Tao, K., Hu, Q., and Shen, Y. (2020).
\newblock Detection of ice hockey players and teams via a two-phase cascaded cnn model.
\newblock {\em IEEE Access}, 8:195062--195073.

\bibitem[He, 2022]{He22}
He, X. (2022).
\newblock {Application of deep learning in video target tracking of soccer players}.
\newblock {\em Soft Computing}, 26(20):10971--10979.

\bibitem[Homayounfar et~al., 2017]{Homayounfar17}
Homayounfar, N., Fidler, S., and Urtasun, R. (2017).
\newblock Sports field localization via deep structured models.
\newblock In {\em 2017 IEEE Conference on Computer Vision and Pattern Recognition (CVPR)}, pages 4012--4020.

\bibitem[Johnson and Everingham, 2010]{Johnson10}
Johnson, S. and Everingham, M. (2010).
\newblock Clustered pose and nonlinear appearance models for human pose estimation.
\newblock In {\em Proc. BMVC}, pages 12.1--11.

\bibitem[Kamble et~al., 2019]{Kamble19}
Kamble, P., Keskar, A., and Bhurchandi, K. (2019).
\newblock A deep learning ball tracking system in soccer videos.
\newblock {\em Opto-Electronics Review}, 27(1):58--69.

\bibitem[Kay et~al., 2017]{Kay17}
Kay, W., Carreira, J., Simonyan, K., Zhang, B., Hillier, C., Vijayanarasimhan, S., Viola, F., Green, T., Back, T., Natsev, P., Suleyman, M., and Zisserman, A. (2017).
\newblock {The Kinetics Human Action Video Dataset}.
\newblock {\em CoRR}.

\bibitem[Lee et~al., 2020]{JungSoo20}
Lee, J., Moon, S., Nam, D.-W., Lee, J., Oh, A.~R., and Yoo, W. (2020).
\newblock A study on sports player tracking based on video using deep learning.
\newblock In {\em 2020 International Conference on Information and Communication Technology Convergence (ICTC)}, pages 1161--1163.

\bibitem[{Li} and {Li Fei-Fei}, 2007]{Li07}
{Li}, L. and {Li Fei-Fei} (2007).
\newblock What, where and who? classifying events by scene and object recognition.
\newblock In {\em 2007 IEEE 11th International Conference on Computer Vision}, pages 1--8.

\bibitem[Li et~al., 2023]{Li23}
Li, L., Zhang, T., Kang, Z., and Zhang, W.-H. (2023).
\newblock Design and implementation of a soccer ball detection system with multiple cameras.
\newblock {\em ArXiv}, abs/2302.00123.

\bibitem[Microsoft, 2023]{microsoftlaliga}
Microsoft (2023).
\newblock Shaping the future of the game.
\newblock Accessed on November 3, 2023.

\bibitem[Parmar and Morris, 2019a]{Parmar19wacv}
Parmar, P. and Morris, B. (2019a).
\newblock Action quality assessment across multiple actions.
\newblock In {\em {IEEE} Winter Conference on Applications of Computer Vision, {WACV} 2019, Waikoloa Village, HI, USA, January 7-11, 2019}, pages 1468--1476. {IEEE}.

\bibitem[Parmar and Morris, 2019b]{Parmar19}
Parmar, P. and Morris, B.~T. (2019b).
\newblock What and how well you performed? {A} multitask learning approach to action quality assessment.
\newblock In {\em {IEEE} Conference on Computer Vision and Pattern Recognition, {CVPR} 2019, Long Beach, CA, USA, June 16-20, 2019}, pages 304--313. Computer Vision Foundation / {IEEE}.

\bibitem[Penate-Sanchez et~al., 2020]{Penate20-prl}
Penate-Sanchez, A., Freire-Obreg\'on, D., Lorenzo-Meli\'an, A., Lorenzo-Navarro, J., and Castrill\'on-Santana, M. (2020).
\newblock {TGC20ReId: A} dataset for sport event re-identification in the wild.
\newblock {\em Pattern Recognition Letters}, 138:355--361.

\bibitem[Santana et~al., 2023]{ojsantana22mtool}
Santana, O.~J., Freire-Obregón, D., Hernández-Sosa, D., Lorenzo-Navarro, J., Sánchez-Nielsen, E., and Castrillón-Santana, M. (2023).
\newblock Facial expression analysis in a wild sporting environment.
\newblock {\em Multimedia Tools and Applications}, 82(8):11395--11415.

\bibitem[Shih, 2018]{Shih18}
Shih, H.-C. (2018).
\newblock A survey of content-aware video analysis for sports.
\newblock {\em IEEE Transactions on Circuits and Systems for Video Technology}, 28(5):1212--1231.

\bibitem[Simonyan and Zisserman, 2014]{C2D14}
Simonyan, K. and Zisserman, A. (2014).
\newblock Two-stream convolutional networks for action recognition in videos.
\newblock {\em ArXiv}, abs/1406.2199.

\bibitem[Stein et~al., 2018]{Stein18}
Stein, M., Janetzko, H., Lamprecht, A., Breitkreutz, T., Zimmermann, P., Goldlücke, B., Schreck, T., Andrienko, G., Grossniklaus, M., and Keim, D.~A. (2018).
\newblock Bring it to the pitch: Combining video and movement data to enhance team sport analysis.
\newblock {\em IEEE Transactions on Visualization and Computer Graphics}, 24(1):13--22.

\bibitem[Teranishi et~al., 2020]{Teranishi20}
Teranishi, M., Fujii, K., and Takeda, K. (2020).
\newblock Trajectory prediction with imitation learning reflecting defensive evaluation in team sports.
\newblock In {\em 2020 IEEE 9th Global Conference on Consumer Electronics (GCCE)}, pages 124--125.

\bibitem[Wang et~al., 2019]{Shaobo19}
Wang, S., Xu, Y., Zheng, Y., Zhu, M., Yao, H., and Xiao, Z. (2019).
\newblock Tracking a golf ball with high-speed stereo vision system.
\newblock {\em IEEE Transactions on Instrumentation and Measurement}, 68(8):2742--2754.

\bibitem[Wang et~al., 2017]{NonlocalNN17}
Wang, X., Girshick, R.~B., Gupta, A.~K., and He, K. (2017).
\newblock Non-local neural networks.
\newblock {\em 2018 IEEE/CVF Conference on Computer Vision and Pattern Recognition}, pages 7794--7803.

\bibitem[Wu et~al., 2019]{Wu19}
Wu, Y., Xie, X., Wang, J., Deng, D., Liang, H., Zhang, H., Cheng, S., and Chen, W. (2019).
\newblock Forvizor: Visualizing spatio-temporal team formations in soccer.
\newblock {\em IEEE Transactions on Visualization and Computer Graphics}, 25(1):65--75.

\bibitem[Xu et~al., 2020]{Xu20}
Xu, C., Fu, Y., Zhang, B., Chen, Z., Jiang, Y.-G., and Xue, X. (2020).
\newblock Learning to score figure skating sport videos.
\newblock {\em IEEE Transactions on Circuits and Systems for Video Technology}, 30(12):4578--4590.

\bibitem[Zhang et~al., 2021]{zhang2021bytetrack}
Zhang, Y., Sun, P., Jiang, Y., Yu, D., Yuan, Z., Luo, P., Liu, W., and Wang, X. (2021).
\newblock {ByteTrack: Multi-Object Tracking by Associating Every Detection Box}.
\newblock In {\em European Conference on Computer Vision}.

\end{thebibliography}
}

\end{document}